# Layer-specific Optimization for Mixed Data Flow with Mixed Precision in FPGA Design for CNN-based Object Detectors

Duy Thanh Nguyen, Hyun Kim, *Member, IEEE*, and Hyuk-Jae Lee, *Member, IEEE*

*Abstract*— **Convolutional neural networks (CNNs) require both intensive computation and frequent memory access, which lead to a low processing speed and large power dissipation. Although the characteristics of the different layers in a CNN are frequently quite different, previous hardware designs have employed common optimization schemes for them. This paper proposes a layer-specific design that employs different organizations that are optimized for the different layers. The proposed design employs two layer-specific optimizations: layer-specific mixed data flow and layer-specific mixed precision. The mixed data flow aims to minimize the off-chip access while demanding a minimal on-chip memory (BRAM) resource of an FPGA device. The mixed precision quantization is to achieve both a lossless accuracy and an aggressive model compression, thereby further reducing the off-chip access. A Bayesian optimization approach is used to select the best sparsity for each layer, achieving the best trade-off between the accuracy and compression. This mixing scheme allows the entire network model to be stored in BRAMs of the FPGA to aggressively reduce the off-chip access, and thereby achieves a significant performance enhancement. The model size is reduced by 22.66-28.93 times compared to that in a full-precision network with a negligible degradation of accuracy on VOC, COCO, and ImageNet datasets. Furthermore, the combination of mixed dataflow and mixed precision significantly outperforms the previous works in terms of both throughput, off-chip access, and on-chip memory requirement.**

## I. INTRODUCTION

In computer vision, object detection is a challenging task. Recently, deep learning has been widely adopted in object detection owing to the support of powerful computation devices

This work was partly supported by Institute of Information & communications Technology Planning & Evaluation (IITP) grant funded by the Korea government(MSIT) (No.2020-0-01305, Development of AI Deep-Learning Processor and Module for 2,000 TFLOPS Server) and the Basic Science Research Program through the National Research Foundation of Korea (NRF) funded by the Ministry of Education under Grant NRF-2019R1A6A1A03032119. *(Corresponding author: H. Kim.)*

D. T. Nguyen, and H.-J. Lee are with the Inter-University Semiconductor Research Center, Department of Electrical and Computer Engineering, Seoul National University, Seoul 08826, Korea (E-mail: {thanhnd, hyuk_jae_lee}@capp.snu.ac.kr).

H. Kim is with the Department of Electrical and Information Engineering and the Research Center for Electrical and Information Technology, Seoul National University of Science and Technology, Seoul 01811, Korea (e-mail: hyunkim@seoultech.ac.kr).

such as a GPU and an FPGA. Therefore, numerous promising approaches have been proposed for object detection with deep learning such as single-shot multibox detection (SSD) [1], faster R-CNN [2], RetinaNet [3], DSSD [4], and YOLO [5]. Among these detectors, for object detection, YOLO performs one of the best trade-offs between accuracy and speed [6]. It is a single neural network that predicts both the object bounding boxes and class probabilities.

For achieving real-time operation, numerous FPGA designs are available for a YOLO CNN [7]-[11]. The previous designs in [7], [10], and [11] achieve a real-time throughput. However, these designs only implement tiny YOLO-v2, which is relatively shallow, and thus, achieving a relatively low detection accuracy. The design in [8] combines a binary network for feature extraction and a support vector machine (SVM). The detection accuracy (mAP) is reported as 67.6%, and the frame rate is 40.8 fps (frame-per-second) for a relatively small input image (i.e., 224×224). For achieving a better throughput and hardware efficiency, the study in [9] presents a streaming design for binary weight YOLO-v2. The data path is optimized to maximize the data reuse and eliminate the off-chip access. Thus, this design realized a high throughput with a minimum DRAM bandwidth, which results in low power consumption in the DRAM access. Similarly, the CNN accelerator design in [12] and [13] also uses a single bit to attain an extremely high compression rate and a low hardware cost. However, binary weight quantization causes a significant accuracy degradation because it ignores the effect of the large weights on the detection accuracy, which is not negligible even though only a few large weights are present.

To avoid an accuracy decrease, several previous designs employ quantization with 8 or 16-bit fixed point numbers for the parameters [14]-[16]. However, owing to the large model size, for its computation, these designs require the parameters of each layer to be loaded from an external memory. Consequently, their throughputs are relatively low and power consumption is high. The design in [17] combines 4-bit weights and a small portion of 8-bit weights to achieve an impressive model compression while realizing a nearly lossless accuracy. For sparse computations, this design may suffer from load imbalance because the partition value is obtained by sorting the weights of the entire layer (i.e., layer-wise). To alleviate this problem, a relatively more complex ASIC design is adopted in [18]. However, this design assumes that throughout a given network, the portion of the outliers is fixed. Therefore, the model size can be further reduced as each layer has a different portion of outliers. Moreover, as this design runs a single layer at a time, it requires an extremely large on-chip SRAM for the intermediate data to reduce the off-chip memory access.

Previous designs for object detection CNNs use a common hardware organization and optimization scheme for all the layers in a CNN. However, the sizes of the feature maps and parameters of the different layers are often quite varied. The parameter sparsity may also change depending on the layer. Therefore, using a common organization for all the layers may not be an effective approach to design a CNN hardware accelerator. In this view, this paper presents a layer-specific design that employs different organizations that are optimized for the different layers. The proposed design employs two layer-specific optimizations: layer-specific mixed precision and layer-specific mixed data flow.

For layer-specific mixed precision, the proposed design uses dense 1-bit weights and sparse 8-bit weights to achieve a nearly lossless accuracy with a significant reduction in the model size. The ratio of the 1-bit and 8-bit weights is chosen carefully to minimize the required data size while avoiding an accuracy decrease. For layer-specific mixed data flow, the hardware organization is selected according to the sizes of the feature maps and parameters. The main contributions of this paper are summarized as follows:

- Algorithmic contribution: A mixed precision quantization with a retraining method is proposed. The Gaussian optimization method is applied to select the best sparsity for the trade-off between compression and accuracy for each layer independently, and consequently, the quantized network outperforms the binary weight network while achieving a similar compression ratio. The proposed scheme causes a negligible accuracy (less than 1%) while reducing the memory size by 22.66 – 28.93 times, compared to a full-precision network.

- Architectural contribution: A mixed precision streaming architecture with a mixed data flow is proposed. Compared to "Shortcut Mining" [19], the mixed data flow scheme reduces the off-chip access for feature-maps from 62MB to 0 MB while requiring smaller BRAM sizes and achieving higher throughput. Compared to the unified design, the combination of the two proposed schemes runs 3.2 times faster, while reducing on-chip memory size by 2.0 times and requiring 12.95 times less off-chip access.

The remained of this paper is organized as follows. Section II discusses about the previous works on CNN hardware design. Section III and IV present the proposed mixed data flow and mixed precision quantization, respectively. In Section V, the hardware architecture for the proposed quantization is elaborated. The experimental results are presented in Section VI. Finally, Section VII concludes the paper.

## II. BACKGROUND AND RELATED WORKS

### A. CNN Data Flows

Fig. 1 presents two main scheduling schemes of weight reuse for tile-based convolution. The sizes of the tiled input channels and output channels are denoted as $T_i$ and $T_o$, respectively. H is the feature map size, and N and M are the numbers of the input channels and output channels, respectively. The advantages and drawbacks of each strategy are described below.

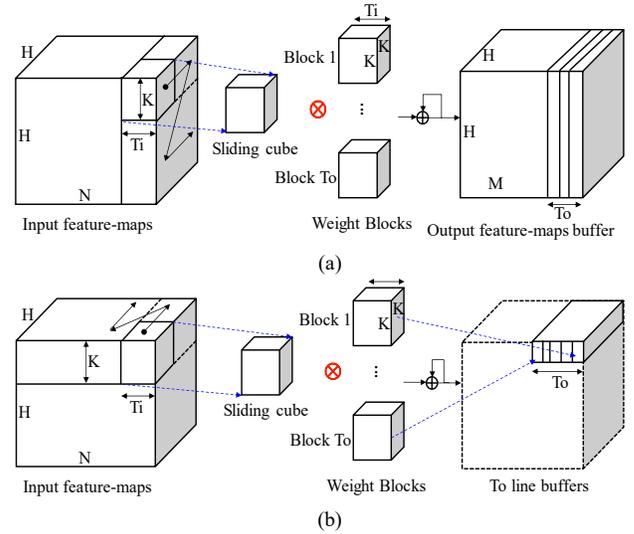

Fig. 1. The scheduling for streaming convolutional layer: (a) Full weight reuse (Scheme 2). (b) Row-based weight reuse (Scheme 3). Borrowed from Fig. 2 in [9].

The "Scheme 2" shown in Fig. 1(a) maximizes weight reuse. Each weight is reused for the whole input channel (i.e., reuse $H^2$ times). At a time, $T_i$ input planes are convolved with each of $T_o$ weight blocks. The partial sum is stored in an output buffer. The SRAM size of this output buffer is $T_o \times H^2 \times Q_S$, where $Q_S$ is the bit width of the accumulation before quantization. To produce final $T_o$ output feature-maps, the entire input feature-maps are accessed. Hence, to generate the entire output feature-maps, the input feature-maps are repeatedly read $M/T_o$ times. Therefore, the input buffers must be large to store them. Moreover, to pipeline between layers, the input buffer size should be doubled, which is $2 \times H^2 \times N \times Q_A$, where $Q_A$ is the bit width of input feature-maps.

The "Scheme 3" shown in Fig. 1(b) minimizes the input buffer size and processing latency. The input sliding cube (i.e., $K \times K \times T_i$ pixels) slides along the width of the input image, which is called a *row pass*. The input sliding cube is convolved with $T_o$ weight blocks each time to produce $T_o$ temporary output values. These weight blocks are reused for a *row pass*. These $T_o$ computations are processed in parallel and saved in the line buffers thereby creating $T_o$ temporary output channels. The input sliding cube then shifts $T_i$ channels toward the end of N-input channels. In the next *row pass*, new $T_o$ weight blocks are fetched and convolved with the new sliding cube. To finish the processing for one line, the entire weight of the model is accessed from the memory. Therefore, to process the whole input feature-maps, the weights are read H times. Regarding the hardware resource, the input buffer size for pipelining is $(K+1) \times N \times H \times Q_A$, and the temporary accumulation buffer size is $T_o \times H \times Q_S$.

### B. Related Works about CNN Accelerators

Deep neural networks such as VGGNet [20], YOLOv3 [21], and ResNet-152 [22] are very powerful. However, they consume a huge amount of memory bandwidth and computational resources. Frequent access to off-chip memory causes long latency and large energy consumption [23], [24]. For example, a 32-bit floating point addition consumes only 1pJ while a 32-bit data word access from DRAM (45nm CMOS technology) requires 640pJ [25].

There are a number of previous works designing a single layer accelerator for CNNs [13]-[15], [16], [18], [26]-[29], [31]. These works optimize the processing of a single CNN layer through loop optimization to increase hardware utilization. However, the performance of the network decreases as it goes deeper owing to data transfer back and forth between the CNN accelerator and off-chip memory. For example, ResNet-152 layers [22] requires 102.8 MB of weights and 61.1 MB of feature-maps of 16-bit precision. The ideal technique would need to load off-chip data once for computation.

To reduce off-chip access, fused layer techniques [32], [33] cascade multiple layers. The intermediate feature-maps pyramid is stored in BRAMs, thereby consuming considerable amount of BRAMs resource (e.g, 5 layers of VGGNet require 122% of the BRAMs in Xilinx Virtex-7 chip). Therefore, this technique does not scale up well for deeper networks owing to their large intermediate feature-maps storage.

There are several works that presents a multi-layer processor approach, in which each layer is processed by a dedicated hardware unit [10], [11], [34], [35], to maximize the utilization of computing resources. As the on-chip memory is not enough for multiple hardware units, the data have to be stored in off-chip memory. Therefore, these works require huge amount of memory access for data. Even they work fine for shallow networks, it is difficult to scale up to deeper networks.

Flexible data flow has been studied in many previous works: Flexflow [36], DNA [37], SmartShuttle [28], and MAERI [38]. Flexflow demonstrates a unified design with a combination of feature-maps, neuron, and synapse-level parallelism to boost the resource utilization. DNA leverages the input, output, and weight reuse within the same fabric. Each single layer is assigned a reuse pattern to achieve the best resource utilization. However, both designs are applied to a single layer, not across different layers. These designs aim to boost the resource utilization, not optimizing the off-chip memory access/on-chip memory size. SmartShuttle allows switching among two data reuse schemes: partial sum reuse oriented and weight reuse oriented by using an empirical method for choosing tiling factors. Running 13 CONV layers of VGGNet with SmartShuttle requires 434.8 MAC/DRAM access (i.e., 142 MB). The off-chip access is projected to be larger for deep network such as ResNet152 because it requires large shortcut outputs and feature-maps. Unlike conventional CNN computation, MAERI has a tree-based reconfigurable interconnect within the accelerator to handle convolution, recurrent layers with irregular filter sizes and sparsity. This work aims to maximize the data mapping to MACs, not the on-chip/off-chip memory access directly.

A recent work [19] presents "Shortcut Mining", an accelerator design with a flexible buffer structure, to maximize the reuse of shortcut feature-maps. It achieves a significant speed up over the previous works owing to the reduction of off-chip access for feature-maps. It is a single layer design with shortcut buffer optimization. Even though it reduces the off-chip access for shortcut data significantly, this work still requires large off-chip access while consuming most of available on-chip memory resource of an FPGA chip.

To aggressively reduce the on-chip/off-chip utilization, this paper proposes a new hardware architecture which has two parts: pipelined layers and main layer. Similar to the fused layer design [32], the functionality of pipelined layers is to reduce off-chip access for feature-maps completely. Unlike the fused technique, which stores 2D feature-maps pyramid of consecutive layers on-chip, this work does pipeline between layers based on line buffers. Therefore, this work consumes much smaller on-chip resources than the design in [32]. The main layer processes the remaining layers while being able to store intermediate data on-chip. Hence, this work completely removes the off-chip access for feature-maps with a small on-chip memory size. Moreover, the mixed precision compression further reduces off-chip accesses and speeds up the computation. The detail of the proposed work is to be discussed in the following sections.

## III. LAYER-SPECIFIC MIXED DATA FLOW DESIGN

### A. CNN Accelerator with the Mixed Data Flow

Fig. 2 illustrates the memory requirements for each layer in Sim-YOLO-v2. In the beginning layers, the feature map sizes are quite large, whereas the parameter sizes are relatively small. For example, CONV1 outputs 5.5 million feature maps while using only 864 parameters. Therefore, the "Scheme 3" is suitable for the beginning layers because it is less demanding on the row buffer. Contrastingly, the last few layers generate small feature maps but require numerous parameters. Therefore, the second scheme outperforms the "Scheme 3" owing to its full reuse of the weight parameters. Accordingly, this paper proposes an accelerator design with a mixed data flow to optimize the data reuse for each layer and reduce the BRAM utilization. As depicted in Fig. 3, the proposed scheme decomposes the network into two groups of layers. For the layers in the first group, each layer is processed by its dedicated hardware unit. For reducing the external memory access, these layers are pipelined using line buffers, and the "Scheme 3" is used for the row-based weight reuse. This implies that each layer starts its computation as soon as several rows of its inputs are delivered. Therefore, the delay between these layers is short. For plain networks such as AlexNet, VGGNet, and SimYOLOv2, each block is considered as a single layer, and there is no need of delayed shortcut buffer. On the other hand, for residual style networks such as ResNet and YOLOv3, each block is a residual block with a delayed shortcut line buffer. The number of line buffers is calculated by the number of delayed lines between the first and last layer in a residual block. For example, in Fig. 3(b), delay between input and output of 3×3 and 1×1 convolutional layer is 2 and 1, respectively. If the output of a block is used for a deep layer later, it is stored to off-chip memory instead of using shortcut line buffer to save on-chip memory. As the parameter sizes of these layers are small, they are stored in an external memory to further save the BRAM resources. This data reuse scheme eliminates numerous DRAM accesses for intermediate data (feature maps) while demanding a few DRAM accesses for the parameters. By contrast, the layers in the second group are processed sequentially using a single hardware unit represented as the "Main Layer" in the Fig. 3. For scheduling, the full weight reuse scheme (i.e., scheme 2) is employed. The intermediate data are small in size, and thereby stored entirely in BRAMs. Moreover, the parameters are read once for each layer so that the DRAM bandwidth is kept small. It should be noted that the main layer has a similar architecture to that of the pipeline layers, except the manner in which the sliding cube moves (i.e., the data flow). It includes an input frame buffer, output frame buffer, and shortcut frame buffer. The size of the shortcut buffer is the maximum size of shortcut's output. These buffers are interchanged for consecutive layers to minimize the intermediate on-chip data movement. For example, beside storing shortcut data, shortcut buffer can function as an input buffer if its data are used for the next layer. To pipeline the processing between two groups, one more input buffer is needed. It is worth mentioning that the proposed design does not require a DRAM access for intermediate data (feature maps). Instead, they are all stored in BRAMs inside an FPGA device.

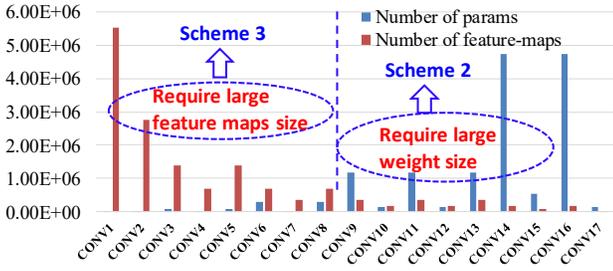

Fig. 2. Memory requirements for each layer in SimYOLOv2. Scheme 2 (full weight reuse) and Scheme 3 (row-based weight reuse) are brought from Table I in [9]. It should be noted that Scheme 1 (no reuse) in [9] is not listed here.

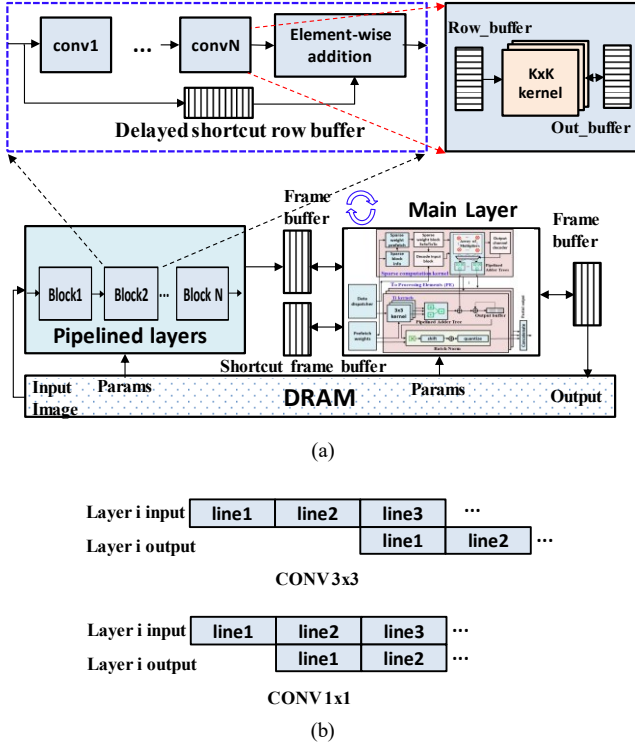

Fig. 3. Accellerator design with mixed data flow. The detailed architecture of "Main Layer" is discussed in section V.

### B. Optimization for the Mixed Data Flow Design

The consequent problem is the decomposition of the network into two groups for the best trade-off between the SRAM size and DRAM bandwidth. In the mathematical formulation, the number of layers in the network and group boundary are denoted as L and i, respectively. The SRAM size and total DRAM access per input image are calculated as a function of i. For simplicity, the DRAM accesses for the input image and final outputs are not presented here.

$$SRAMsize(i) = \sum_{l=1}^{i}(row\_buff_l + out\_buff_l) + 3 \times \max_{l \in [i+1,L]}(in\_frame\_buff_l) + \max_{l \in [i+1,L]}(shortcut\_buff_l) + \max_{l \in [i+1,L]}(out\_buff_l) \quad (1)$$

where $row\_buff_l$ and $out\_buff_l$ are respectively the input buffer and partial sum buffer for layer $l$ in the first group, and $in\_frame\_buff_l$ is the input frame buffer for layer $l$ in the second group. $\max_{l \in [i+1,L]}(shortcut\_buff_l)$ is the maximum size of shortcut's output in the second group. As shown in Table I in [9], for layers in [1,i], $row\_buff_l = (K_l+1) \times N_l \times H_l \times Q_A$, $out\_buff_l = T_o \times H_l \times Q_S$, and for layers in [i+1,L], $in\_frame\_buff_l = H_l^2 \times N_l \times Q_A$, $out\_buff_l = T_o \times H_l^2 \times Q_S$, respectively. $K_l, H_l, N_l, M_l$ are the kernel size, feature map width (height), number of input channels, and number of output channels, respectively. $Q_A$, $Q_S$ are the bit-width of input activation, and partial sum. $T_i$ and $T_o$ are the tiling factors.

$$DRAMaccess(i) = \sum_{l=1}^{i} H_l \times param_l + \sum_{l=i+1}^{L} param_l \quad (2)$$

where $param_l$ is the parameter size of layer $l$.

If the parameters of the first group are stored in SRAMs, the SRAM size and DRAM access with respect to boundary $i$ are formulated as equations (3) and (4), respectively. Compared to (2), the DRAM access is reduced by $\sum_{l=1}^{i} H_l \times param_l$.

$$SRAMsize(i) = \sum_{l=1}^{i}(row\_buff_l + out\_buff_l + param_l) + 3 \times \max_{l \in [i+1,L]}(in\_frame\_buff_l) + \max_{l \in [i+1,L]}(shortcut\_buff_l) + \max_{l \in [i+1,L]}(out\_buff_l) \quad (3)$$

$$DRAMaccess(i) = \sum_{l=i+1}^{L} param_l \quad (4)$$

In addition to the SRAM size reduction, the throughput is also an important factor in the selection of the group boundary. To achieve a balanced pipeline between two groups, the computation times of the two groups need to be similar. Owing to the fully pipelined design of the convolutional layers, their outputs are obtained in each cycle. It is assumed that the provided DRAM bandwidth is sufficient to not affect the execution time. In fact, this assumption is true in most design options considered in this study. The computation time of layer l is calculated as follows:

$$t_l = \frac{MACS_l}{K_l^2 PF_l} = \frac{H_l^2 N_l M_l}{PF_l} \quad (5)$$

where $MACS_l$ represents the number of MAC operations in the lth layer. $PF_l = T_i(l) \times T_o(l)$ is the parallelism factor, which is proportional to the number of multipliers in lth layer ($K_l^2 PF_l$). It is noteworthy that for the layers in the first groups, for a certain integer, $X$, $PF_l = 2^X$ for a certain integer. Concurrently, in the main layer, for a certain integer $Y$, $PF_l = 2^{2Y}$ for a certain integer owing to the power-of-two tiling factors ($T_i = T_o$). The output buffer of the current layer is the input buffer of the next layer (i.e., $T_i = T_o$) to simplify the control logic and unified SRAM bit-width for the frame buffers.

To balance the pipeline between consecutive layers in the first group, the parallelism factors in the first group must satisfy the condition: $t_1 = t_2 = \cdots = t_i$. It is noteworthy that $T_i(l+1) = T_o(l)$ and $M_l = N_{l+1}$. Hence, $T_i$ and $T_o$ of each layer can be easily chosen by the guideline in (6).

$$\begin{cases} T_i(l+1) = T_o(l) \\ T_o(l+1) = \frac{H_{l+1}^2 M_{l+1} T_i(l)}{H_l^2 N_l} \end{cases} \quad (6)$$

The computation times of the first and second groups are calculated as (7) and (8).

$$t_{g1} = \sum_{l=1}^{i-1} D_l \frac{t_l}{H_l} + t_i \quad (7)$$

$$t_{g2} = \sum_{l=i+1}^{L} t_l \quad (8)$$

where $D_l$ is the number of the delayed rows from layer $l$ to $l+1$ in group 1 owing to the pipeline.

For a given set of parallelism factors, group boundary $i$, which ensures a balanced pipeline between the first and second group, is chosen such that:

$$i = \underset{[1,L]}{\operatorname{argmin}} \left| \sum_{l=1}^{i-1} D_l \frac{t_l}{C_l} + t_i - \sum_{l=i+1}^{L} t_l \right| \quad (9)$$

For the mixed data flow design, Algorithm 1 describes the procedure to select the parallelism factors. It is noteworthy that to obtain a solution of Algorithm 1, SRAM constraints $\alpha$ cannot be arbitrarily small for a given network.

## IV. LAYER-SPECIFIC MIXED PRECISION TRAINING

### A. Motivation of Intra-layer Mixed Precision Training

Fig. 4 shows the weight histogram of a channel in the fourteenth convolution layer of Sim-YOLO-v2 [9]. As shown in Fig. 4(a), most weights have small absolute values (e.g, less than 0.02), whereas a few weights have large values. Fig. 4(b) exhibits the binary weight quantization [39]. Both small and large weights are represented by their mean values and signs, respectively. This quantization aggressively reduces the model size while losing numerous weight levels.

**Algorithm 1: Parallelism factors for the proposed scheme.**

**For group boundary $i = 1:L$ do**
   1. **if** SRAMsize(i) > $\alpha$ (MB)
      **Continue;**
   2. Choose $T_i$, $T_o$ of the $1^{st}$ group by using (6)
   3. Choose $T_i$, $T_o$ of the $2^{nd}$ group such that |tg1-tg2| is
      minimum
   4. **if** total required DSPs > total DSPs of FPGA
      Reduce $T_i$, $T_o$ of the $1^{st}$ group; **go to step 2.**
   5. Estimate frame_rate(i)=1/max(tg1, tg2)
**End_for**
$i = argmax(frame\_rate(i))$

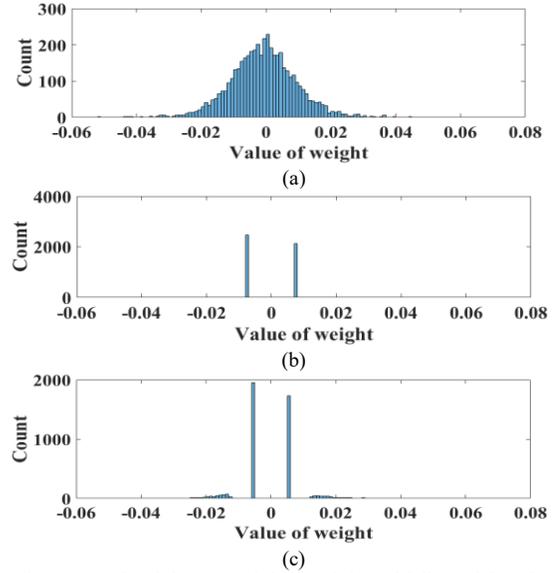

Fig. 4. Histogram of weights (a) Original weights of full precision (b) Binary weight quantization (c) Mixed precision quantization.

To address this problem and achieve a high accuracy while utilizing the advantages of a binary weight quantization, this paper proposes a mixed quantization method by adding more quantization levels, as shown in Fig. 4(c). The weight histogram in Fig. 4(a) is divided into two regions with small and large values. Small weights are quantized using a single bit, whereas a few large weights are quantized in high precision. It should be noted that many previous works such as [40], [55], and [56] show that 8-bit precision is sufficient to achieve near-lossless accuracy for various DNN inferences. In addition, 8-bit precision is also adopted by Google TPU [57]. On the other hand, lower precision, such as 6-bit in [9], causes the significant accuracy loss of 1.23% in the VOC dataset, and consequently, it is not justified to work well for large-scale datasets such as ImageNet. Therefore, in this study, the sparse high-precision weights are quantized to 8 bits to better compensate for the loss caused by binary quantization. Besides, there could be a mixed precision quantization scheme using a gradation of bits from 1 bit to 8 bits. This scheme is plausible for inter-layer approach, in which each layer uses same number of bits [41], [42]. However, applying weights within a same layer is complicated to quantize and inefficient in terms of hardware utilization. Thereby, the proposed scheme utilizes a combination of 1-bit and 8-bits data for intra-layer weight quantization. It is noteworthy that this mixed precision quantization is different from the pruning method in [25] such that it keeps both small weights and large weights. Hence, the proposed method achieves a better approximation to the original weights.

Compared to the outlier-aware scheme in [17], which proposes a combined quantization, the main difference of the proposed scheme is that the binary weight requires no multiplication with the information loss of 1-bit quantization being compensated by a few high-precision weights. Hence, the proposed scheme requires low FPGA resource utilization. Finally, the fixed outlier ratio throughout the network results in a non-optimized model compression. The next subsection and Section V of this paper present the technique to complement these problems.

### B. Coarse-grained Intra-layer Mixed Precision Quantization

In mixed precision quantization, the large weights are divided into two parts, as illustrated in Fig. 5. One part is "+*mean*" value, which is represented as a dense binary weight filter. The other is the original large value subtracted by the "*mean*" value. This scheme enables to design a simple mixed architecture. The proposed design includes two parts: a dense binary convolutional kernel and a sparse 8-bit convolutional kernel. The dense convolutional kernel design is the same as that in a preceding research in [9], which requires no multiplication. The remaining work is only a design of a sparse computation kernel.

A sparse computation inherently causes a load unbalance. Different from [17], this work sorts the weights in the same filter (i.e., channel-wise) to achieve a better computation balance. To accelerate the training, this study uses a segmented sort algorithm, which is supported by the *thrust* library in CUDA. As shown in Fig. 5, the sorted weights of each output kernel are partitioned into small weights and large weights. The major portions are quantized to a single bit, whereas the minor large weights are quantized to eight bits. Each layer is quantized independently to choose the best sparsity for the trade-offs between the accuracy and compression ratio (i.e., hardware cost). Other layers are initialized with a pre-trained full precision model, and they perform only forward computation. The fine-tuning updates only the layer to be optimized. This problem is formulated as an optimization problem shown below:

$$p_i = argmax(L(p_i)) \qquad (10)$$

where $p_i$ is the sparsity of the $i$-th layer, and the objective function is expressed as follows:

$$L(p_i) = mAP(p_i) + \gamma \times C(p_i) \qquad (11)$$

where $mAP(p_i)$ is the mean average precision of the network with respect to the high-precision ratio, $p_i$, of the $i$th layer. Compression rate $C(p_i)$ of this layer is computed as follows:

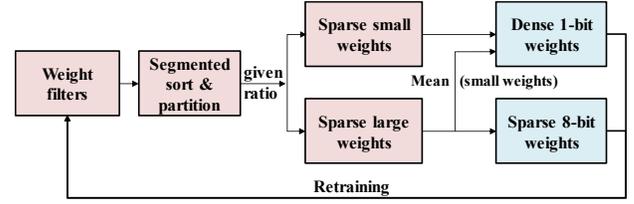

Fig. 5. Block diagram of mixed precision quantization.

$$C(p_i) = \frac{32}{1\times(1-p_i)+8\times p_i} \times \frac{N_i}{\sum N_i} = \frac{32}{1+7\times p_i} \times \frac{N_i}{\sum N_i} \qquad (12)$$

where $N_i$ and $\sum N_i$ are the number of parameters of the $i$th layer and total number of parameters in this network, respectively. $\gamma$ is the coefficient to balance the accuracy and compression benefits.

The relationship between the objective function and variable is not explicitly expressed in a closed form. Bayesian optimization provides a general framework for the global optimization of the black box functions that do not require derivatives [43]. As the objective function is unknown, the Bayesian method treats it as a random function. The Gaussian process regression is a powerful and non-parametric Bayesian method to estimate the objective function in exploration and exploitation scenarios [44], [45]. Therefore, the objective function in (10) is modeled as a Gaussian process.

Algorithm 2 describes the general GP-solution for the proposed optimization problem. From a set of previous samples from the system (i.e., accuracy at a given sparsity), the posterior distribution over function $L(p_i)$ is derived as (13) and (14), where $x$ is $p_i$ and $y$ is $L(p_i)$ at a sample point. The mean of the function and the kernel are as follows:

$$m_t(\mathrm{x}) = K(x, \boldsymbol{X_t})[K(\boldsymbol{X_t}, \boldsymbol{X_t}) + \sigma_\epsilon^2 \boldsymbol{I}]^{-1}\boldsymbol{y_t} \qquad (13)$$

$$k_t(x, x') = k(x, x') - K(x, \boldsymbol{X_t})[K(\boldsymbol{X_t}, \boldsymbol{X_t}) + \sigma_\epsilon^2 \boldsymbol{I}]^{-1}K(\boldsymbol{X_t}, x') \qquad (14)$$

where $\boldsymbol{X_t}$ and $\boldsymbol{y_t}$ are the inputs and outputs of the sampled points so far. The acquisition function, $V_t(x)$, is chosen from the *upper confidence bound (UCB)* algorithm [25], which is expressed as follows:

$$V_t(x) = m_{t-1}(x) + \omega_t\sqrt{s_{t-1}(x)} \qquad (15)$$

where $\sqrt{s_{t-1}(x)}$ is the predictive standard deviation at a point, $x$, and $s_t(x) = k_t(x,x)$. $\omega_t$ is a free constant parameter, which performs the trade-off between the expectation and uncertainty. After numerous iterations, the algorithm converges to an optimal value.

It is worth mentioning that each layer behaves differently at the same level of quantization. For example, convolution layers with a large kernel size are more error tolerant to a low-bit quantization than those with a small kernel size [46]. Consequently, large convolution layers can be further compressed with the proposed method owing to their high sparsity. Therefore, the proposed algorithm compresses the

---

**Algorithm 2: Gaussian process (GP) optimization solution.**

**Problem:** *written in (10)*
**Input:** *input space [0,1]; acquisition function $V_t$, GP-prior for $L(p)$ with mean function $m(p)$ and kernel $k(p, p)$*

**For i = 1:N do**
　*Update the posterior of GP and $V_t$ as shown in (13), (14), (15)*
　*Choose $P_t^* = argmax_{p \in [0,1]}(V_t(p))$*
　*Sample $L(P_t^*)$ from system*
**End_for**

---

network efficiently by varying the sparsity of the different layers.

## V. THE HARDWARE ARCHITECTURE WITH MIXED PRECISION

The design of the proposed design is depicted in Fig. 6. The convolutional layer requires additional kernels for sparse computation in parallel to dense one 1-bit computational kernels, as shown in Fig. 6(a). Each output channel has different numbers of sparse weights. It causes the unbalanced computation between output channels if they are processed separately. To solve this imbalance and utilize the multipliers better, all the sparse weights of $K \times K \times T_i \times T_o$ weight blocks are merged and computed simultaneously as proposed in [26]. Fig. 6(b) illustrates the sparse weight format for an entire layer. The memory buffer, named "*Sparse block info,*" stores the number of sparse weights of $K \times K \times T_i \times T_o$ weight blocks, whereas the weight buffer, named "*Sparse weight block,*" stores the sparse weight values and their relative coordinates. The memory requirement for the sparse weights reduces to $O(2.5 \times nonzeros + M)$, where M is the number of $K \times K \times T_i \times T_o$ weight blocks in this layer. This is the same memory efficiency as that of the compressed sparse column (CSC) format proposed in [47]. It is noteworthy that the sparse weights are prefetched in synchronization to the dense weights. If the number of sparse weights for a block is zero, the prefetcher skips reading the sparse weights from the weight buffer, and the sparse computation kernel is also turned off. The sparse weight block is loaded to a buffer. The corresponding activations in the sliding windows are then selected by decoding the coordinates of the sparse weights. The sparse weights and activations are multiplied by an array of multipliers. The number of allocated multipliers for each layer is calculated at the training time. It is the maximum number of sparse weights in sparse weight blocks. In case of the main layer, to support different high precision weight ratio, the number of allocated multipliers is the maximum number of multipliers of layers run by the main layer. Owing to the high sparsity, the size of the multiplier array is small, thereby keeping the hardware overhead of a sparse kernel small. According to the coordinates of the sparse weights, the outputs from multipliers are then input to the corresponding pipelined adder tree as illustrated in Fig. 6(c). In the next step, the outputs from the pipelined adder trees are added to the outputs from the corresponding dense computation kernels. Finally, the accumulated results are written to the output buffer. There are two key parameters in the sparse computation kernel: *N_multipliers* and *tree_size*. *N_multipliers* is the maximum number of sparse weights in $K \times K \times T_i \times T_o$ weight blocks. *tree_size* is the maximum number of sparse weights in $K \times K \times T_i$ weight blocks.

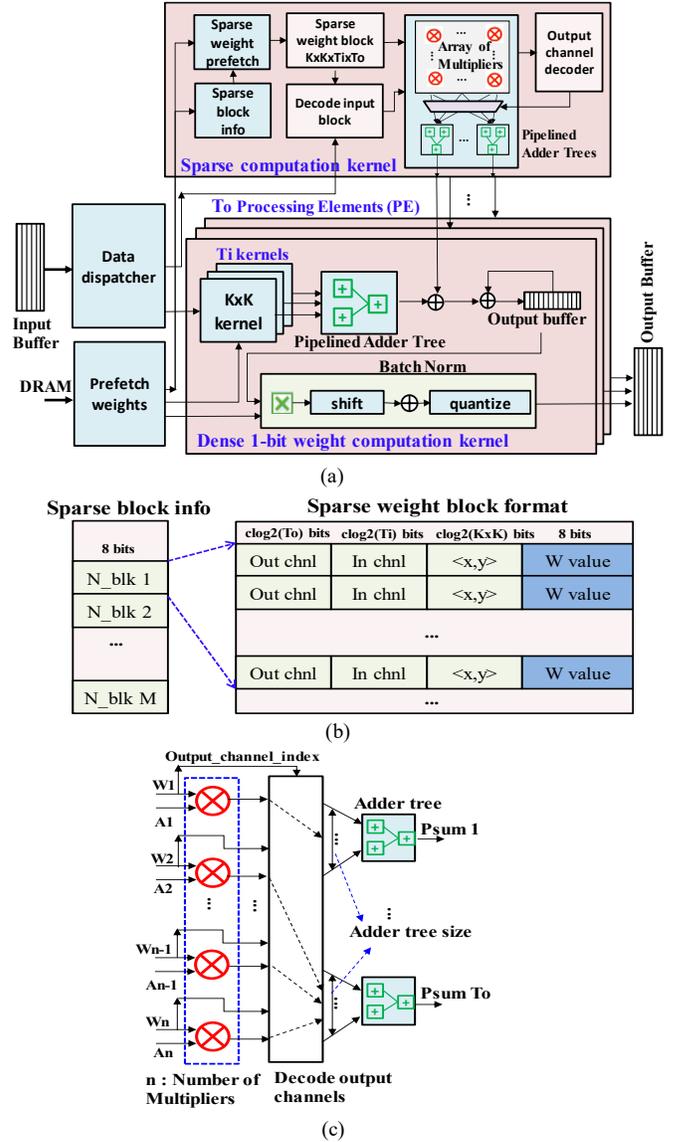

Fig. 6. The proposed mixed precision design. (a) The architecture of a convolutional layer. (b) The sparse weight format for weight blocks of the entire layer. (c) The sparse computation kernel.

Unlike [18] which processes outlier weights and normal weights sequentially, the proposed hardware accelerator computes sparse 8-bits and dense 1-bit kernels in parallel. Both the sparse and dense kernels use the same sliding windows and produce their partial sums in each cycle owing to the fully pipelined design. According to the delay of their pipelined adder tree outputs, some delayed registers need to be added to synchronize the sparse and dense kernels for each sliding windows processing. Owing to this straight synchronization, these two hardware units share the same input buffer, output buffer. The hardware overhead of the proposed mixed design is for the sparse weight buffer, sparse input, multipliers, and pipelined adder tree.

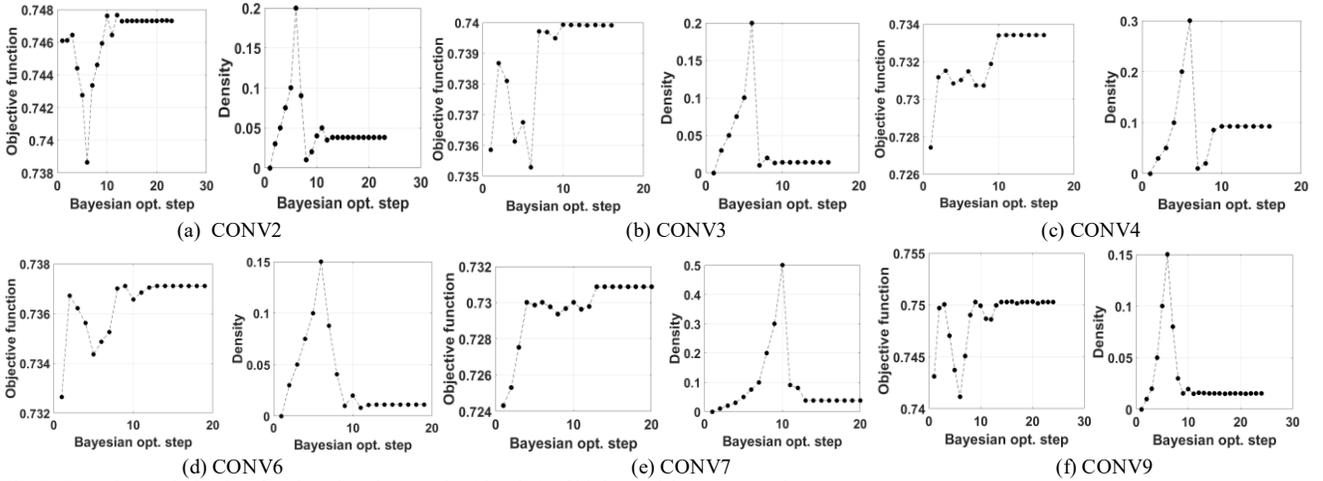

Fig. 7. Bayesian optimization for choosing the sparsity (density) of high-precision computation.

## VI. Experimental Results

### A. Mixed Precision Quantization

In the first experiment, the effectiveness of the Gaussian process optimization, as described in Section IV, is evaluated by compressing Sim-YOLO-v2. Fig. 7 presents the iteration results to choose the best high-precision weight ratio (i.e., density) for some layers. The layers of the same dimension are optimized at once (e.g., CONV3 and CONV5, CONV6, and CONV8). In the proposed work, the balance coefficient, $\gamma$, is set as 0.01. For example, CONV2 converges after at most 15 iterations. The converged ratio (density) is 0.035 (i.e., 3.5% of high-precision weights). As presented in Fig. 7, the optimization for all the layers converges to a small ratio value.

Table I presents the performance of the proposed optimization method, manually chosen methods, and pruning method [25]. It should be noted that the first and last layers are not quantized to preserve the accuracy. Therefore, the high-precision weight ratio of each layer is 1. In the binary quantization, the ratio is set as 0 (i.e., there is no high-precision weight). The compression rate is expressed by the average numbers of bits to represent a parameter. "Mixed ratio 1," and "Mixed ratio 2," are the manually chosen methods. For example, the "Mixed ratio 1" scheme sets all the layer ratios as 0.05. This scheme needs only 1.408 bits, on average, to represent a parameter. Compared to the manual methods, the proposed scheme compresses the network more efficiently using the various ratios optimized for the different layers. Thus, the proposed method realizes a higher accuracy than the manually chosen method with the best compression ratio, "Mixed ratio 2," while obtaining a higher compression ratio. The ratio for some large layers is reduced to less than 1%, such as CONV14 and CONV16. Compared to the full-precision network, the proposed scheme causes a small loss in accuracy (i.e., 0.95%) while requiring a 27.87 times smaller memory size. An additional experimental result for YOLO-v2 tiny is presented in Table II. Compared to the full-precision model, the mixed precision scheme achieves a +0.48% higher accuracy and 28.93 times better compression ratio.

TABLE I
ACCURACY OF MIXED PRECISION QUANTIZATION WITH VARYING HIGH PRECISION WEIGHT RATIO

| Layer | Full precision (ratio = 1) | Binary precision (ratio = 0) | Pruning (ratio of non-zero) | Mixed Ratio 1 | Mixed Ratio 2 | Proposed |
|---|---|---|---|---|---|---|
| Conv1 | 1 | 1 | 1 | 1 | 1 | 1 |
| Conv2 | 1 | 0 | 0.191 | 0.05 | 0.03 | 0.035 |
| Conv3 | 1 | 0 | 0.227 | 0.05 | 0.03 | 0.014 |
| Conv4 | 1 | 0 | 0.178 | 0.05 | 0.03 | 0.093 |
| Conv5 | 1 | 0 | 0.299 | 0.05 | 0.03 | 0.014 |
| Conv6 | 1 | 0 | 0.277 | 0.05 | 0.03 | 0.0112 |
| Conv7 | 1 | 0 | 0.374 | 0.05 | 0.03 | 0.0381 |
| Conv8 | 1 | 0 | 0.325 | 0.05 | 0.03 | 0.0112 |
| Conv9 | 1 | 0 | 0.288 | 0.05 | 0.03 | 0.0154 |
| Conv10 | 1 | 0 | 0.340 | 0.05 | 0.03 | 0.0808 |
| Conv11 | 1 | 0 | 0.233 | 0.05 | 0.03 | 0.0154 |
| Conv12 | 1 | 0 | 0.486 | 0.05 | 0.03 | 0.0808 |
| Conv13 | 1 | 0 | 0.280 | 0.05 | 0.03 | 0.0154 |
| Conv14 | 1 | 0 | 0.431 | 0.05 | 0.03 | 0.0063 |
| Conv15 | 1 | 0 | 0.464 | 0.05 | 0.03 | 0.0663 |
| Conv16 | 1 | 0 | 0.280 | 0.05 | 0.03 | 0.0063 |
| Conv17 | 1 | 1 | 1 | 1 | 1 | 1 |
| **mAP** | 72.08% | 64.95% | 71.00% | 71.68% | 71.02% | 71.13% |
| **# of bits** | 32 | 1 | 2.732 | 1.408 | 1.270 | 1.148 |

TABLE II
COMPARISON OF THE COMPRESSION ON TINY-YOLOV2

| | Original | 1-bit | Mixed precision |
|---|---|---|---|
| **mAP (%)** | 53.96 | 51.44 | 54.44 |
| **# of bits** | 32 | 1 | 1.106 |

For a reasonable comparison to sparse model, the full-precision Sim-YOLO-v2 [9] is pruned and then retrained to obtain the same accuracy as that of the proposed scheme (i.e., within 1% of loss). The non-pruned weights are uniformly quantized to 8 bits. The forth column of the Table I describes the density (non-zero ratio) of the pruned model. Some layers such as CONV12, CONV14, and CONV15 have a density higher than 40%. It is noteworthy that this high density causes the design of a sparse CNN accelerator to be less efficient. Compared to the pruning scheme, the proposed scheme requires a significantly lower high-precision weight ratio for all the layers, thereby yielding a better compression ratio.

To better understand the trade-off relationship between accuracy and model size (i.e., compression rate), various gamma values are selected for the additional experiment. Fig. 8 illustrates the trade-off relationship between accuracy and model size according to various gamma values. When the gamma value is large, the quantization is near to binary quantization to achieve the highest compression rate. On the other hand, the smaller gamma (i.e., less than 0.01) guarantees higher accuracy while sacrificing the compression rate. This result shows the tendency of (11) well, and it is necessary to set a well-balanced gamma value based on this result.

To demonstrate the scalability of the mixed precision scheme to larger networks and datasets, YOLOv3 (106 layers, 65 GOPs) is trained with COCO dataset [48]. All of the original model (i.e., 32-bit floating point), mixed precision, and merging modules are trained for performance comparison. Table III shows that the mixed precision model requires 22.66 times smaller size while losing 0.46% of mAP. Compared to the pruning method with a same compression ratio, the mixed precision scheme achieves 2.7% higher mAP.

Finally, additional tests on ImageNet dataset [49] are conducted to demonstrate the efficiency of the proposed mixed precision scheme. Firstly, the same compressed model of SimYOLOv2 is trained on ImageNet dataset. In Table IV, the proposed scheme reduces model size by 27.37 times while achieves a top-1 accuracy of 72.8% (i.e., 0.1% lower than the floating-point model). Furthermore, Table V shows the comparison of the proposed scheme to the previous works on ResNet-50. HAQ [41] and HAWQ [42] propose mixed precision quantization schemes, in which each entire layer is quantized to a different number of bits. However, they do not cover binary quantization due to a significant loss of accuracy. Whereas, in the proposed work, the accuracy loss due to binary quantization is well compensated by a small dynamic number of high precision weights. Therefore, the proposed work yields a higher performance in terms of both compression efficiency and accuracy.

The above experiments show that even though this paper focuses mainly on YOLO detectors which achieve the best trade-offs between accuracy and speed [6], the proposed scheme also works well for various networks (i.e., YOLOv2, YOLOv3, ResNet50) on various datasets, including both object detection (i.e., VOC, COCO) and classification (i.e., ImageNet). Therefore, it is expected that the proposed scheme can be generalized to all other deep networks based on the convolutional layers.

## B. Accelerator Design with Mixed Precision

The first experiments show the advantages of the proposed channel-wise ratio scheme over the layer-wise scheme proposed in [17]. A large convolutional layer (i.e., CONV14) is compared. Fig. 9(a) depicts the size of the adder tree in each PE (i.e., corresponding to each output channel processing) for both layer-wise and channel-wise mixed precision training. The adder tree size of a layer is the maximum size of the adder tree in all the PEs. Compared to the proposed scheme (i.e., channel-wise), the layer-wise scheme requires a larger adder tree and causes a larger load unbalance between the output channels. Consequently, it demands a relatively larger hardware resource and lower resource utilization. For both the schemes, Fig. 9(b) and 9(c) present the number of multipliers and adder tree size, respectively, for each layer in Sim-YOLO-v2. In Fig. 9(b), the number of multipliers for layers 2, 3, 5, 11, 14, and 16 of the channel-wise scheme is significantly smaller than for those of the layer-wise scheme. The (*mean, standard devi*ation) values of the number of multipliers for the channel-wise and layer-wise schemes are (19.4, 5.0) and (24.7, 7.7), respectively. This indicates that the sparse weights are more uniformly distributed over the weight blocks in the channel-wise scheme than over the layer-wise scheme. A small number of multipliers results in a small LUTs size and low DSP utilization. Fig. 9(c) shows a similar observation that the adder tree size in the channel-wise scheme is many-folds smaller than that in the layer-wise scheme. The (*mean, standard devi*ation) values of the adder tree size are (6.6, 1.7) and (11.1, 4.7) for the channel-wise and layer-wise schemes, respectively. Both the experimental results exhibit that the channel-wise scheme requires relatively much smaller hardware resources and consequently, the hardware utilization is much higher for different layers.

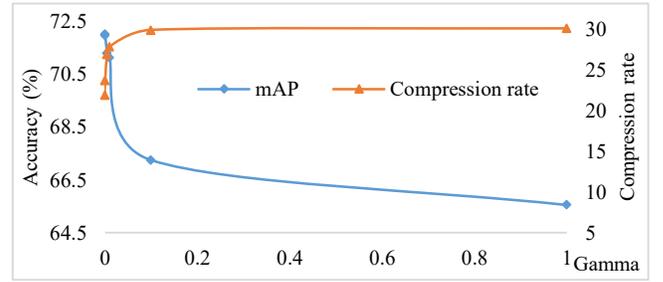

Fig. 8. Trade-off relationship between the accuracy and compression rate.

TABLE III
COMPARISON OF THE COMPRESSION ON YOLO V3 AND COCO DATASETS

| Scheme | Original | Pruning | Mixed precision |
|---|---|---|---|
| mAP (%) | 52.79 | 49.63 | 52.33 |
| # of bits | 32 | 1.412 | 1.412 |

TABLE IV
THE MIXED PRECISION MODEL OF SIMYOLOV2 ON TWO DATASETS

| Scheme | Original | VOC | ImageNet |
|---|---|---|---|
| Accuracy | - | -0.95% (71.13%) | -0.1% (72.80%) |
| Ratio | 1× | 27.87× | 27.37× |

TABLE V
COMPARISON WITH PREVIOUS WORKS FOR THE COMPRESSION ON RESNET-50 AND IMAGENET

| Scheme | Original | HAWQ[42] | HAQ [41] | Mixed precision |
|---|---|---|---|---|
| Top-1 | 75.8% | -0.32% | -0.5% | -0.54% |
| Ratio | 1× | 12.28× | 10.57× | 23.63× |

Table VI provides the comparison of the streaming mixed precision design and previous designs for the YOLO hardware implementation. Owing to the elimination of the off-chip access for intermediate data and parameters, the proposed design attains the same throughput as the precedent design in [9]. In

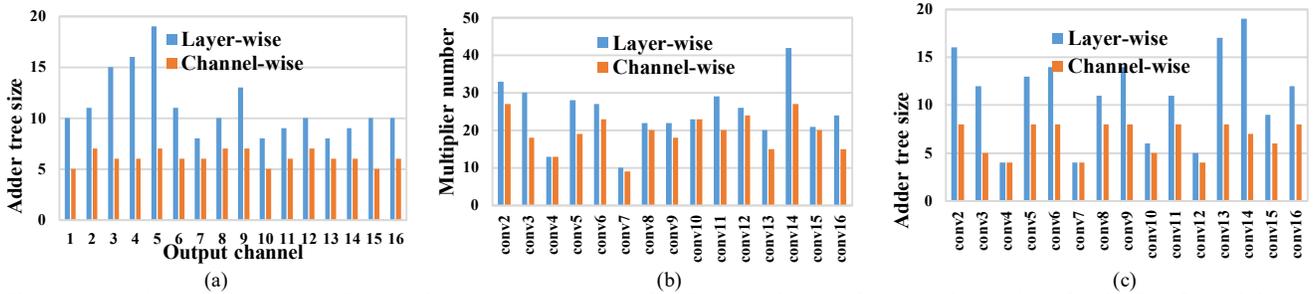

Fig. 9. Comparison of layer-wise vs channel-wise. (a) Adder tree size of each output channel of conv14. (b) Number of multipliers for each layer of SimYOLOv2. (c) Adder tree size of each layer of SimYOLOv2.

TABLE VI
COMPARISON OF THE PROPOSED DESIGN WITH THE PREVIOUS WORKS FOR YOLO CNN HARDWARE

|  | Sim-YOLO-v2 on GPU [5] | Lightweight YOLO-v2 [8] | Tiny-YOLO-v2 [11] | Tiny-YOLO-v2 [9] | Sim-YOLO-v2 on FPGA [9] | This work |
|---|---|---|---|---|---|---|
| Platform | GTX Titan X (16nm) | Zynq Ultrascale+ (16 nm) | Zynq 7000 ZC706 (28 nm) | Virtex-7 VC707 (28 nm) | Virtex-7 VC707 (28 nm) | Virtex-7 VC707 (28 nm) |
| Frequency | 1 GHz | 300 MHz | 200 MHz | 200 MHz | 200 MHz | 200 MHz |
| BRAMs (18 Kb) | N/A | 1706 | 666 | 1026 | 1144 (55.5%(*)) | 1245 (60.4%(*)) |
| DSPs | N/A | 377 | 680 | 168 | 272 (9.7%(*)) | 829 (29.6%(*)) |
| LUTs (k) | N/A | 135 | 86.1 | 86 | 155 (51.5%(*)) | 245.3 (80.8%(*)) |
| CNN Size (GOP) | 17.18 | 14.97 | 10.9 | 6.97 | 17.18 | 17.18 |
| Precision (W, A)(**) | (32, 32) | (1-32, 1-32) | (8, 8) | (1,6) | (1, 6) | Mixed (1,8) |
| Image Size | 416×416 | 224×224 | 1280×384 | 416×416 | 416×416 | 416×416 |
| Frame rate | 88 | 40.81 | 44.2 | 66.56 | 109.3 | 109.3 |
| Accuracy (mAP) (%) | 72.08 | 67.6 | N/A (***) | 51.38 | 64.16 | 71.13 |
| Throughput (GOPS) | 1512 | 610.9 | 468 | 464.7 | 1877 | 1877 |
| DRAM BW (MB/s) | N/A | N/A | 12524 | 52 | 85 | 85 |

Note: (*): % of HW utilization within the FPGA, (**): W: Weight, A: Activation, (***): mAP for only 3 high accuracy classes: car, pedestrian, and cyclist.

terms of the hardware utilization, compared to the precedent design [9] in the sixth column, the proposed design requires only 4.9% more BRAMs utilization and 19.9% more DSPs utilization. In terms of the LUTs, the proposed scheme utilizes 29.3% more LUTs of the FPGA resources. It is noteworthy that mAP of the proposed scheme is 71.13%, which is significantly higher than in [9]. Compared to full-precision Sim-YOLO-v2 on a GPU, in the second column, the proposed accelerator loses only 0.95% detection accuracy while requiring a 27.87 times smaller parameter size, which results in a much lower power consumption. Furthermore, in terms of the throughput, the proposed design achieves a 1.24 times higher performance.

### C. Accelerator Design with Mixed Data Flow

The experimental results show that the hardware accelerator with the mixed data flow is highly efficient in reducing the SRAM size while minimizing the off-chip access.

The first experiment is conducted with 8-bit precision Sim-YOLO-v2. Fig. 10(a) shows the on-chip size and off-chip accesses of the mixed data flow scheme with respect to the group boundary. The "gray" bar graph shows the estimated SRAM size when the parameters of the first group are stored in SRAM. The "DRAM accesses (SRAM)" bars present the off-chip access when the parameters of the first group are stored in SRAM. The "DRAM accesses (DRAM)" line depicts the off-chip access when the parameters of the first group are stored in DRAM. For the case when the parameters in the first group are stored in DRAM, the total DRAM access for a single input image is 191 MB, which is extremely large for high-speed processing on an embedded FPGA. In fact, this case corresponds to the previous streaming design in [9] (i.e., group boundary is CONV17 in Fig. 10(a)). By contrast, the proposed mixed data flow with the group boundary at CONV7 reduces the SRAM size by 8.9 times when compared to the previous streaming design [9]. The DRAM access is 14 MB, which corresponds to a reduction by 13.5 times.

The second experiment shows the advantages of the proposed mixed data flow compared to the state-of-the-art design, *DNNBuilder*, in [11]. For a reasonable comparison, the same YOLO network, data precision (i.e., 16 bits), and input HD image size (1280 × 384) are used. The mixed data flow divides YOLO into two groups. Parameters in both the groups are loaded from a DRAM. Fig. 10(b) illustrates the SRAM size and DRAM access (per single input) with respect to the group boundary. The boundary at CONV4 performs the best in terms of both the factors. However, *DNNBuilder* has all layers only in group 1, implying that the boundary is at the extreme last layer. The required DRAM bandwidth is 10.3 times smaller than that in the design in [11]. Table VII provides the performance comparison of the proposed work and result given in [11] on a ZC706 FPGA board. While achieving the same throughput, the mixed data flow scheme requires 1.61 times less BRAMs and a 10.3 times lower DRAM bandwidth.

To demonstrate the scalability of the proposed mixed data flow design to more complicated networks, two very deep networks, ResNet-152 with 152 layers and YOLOv3 with 106 layers, are deployed in the proposed hardware design. It is assumed that the parameters of the first group are stored in the on-chip memory. Fig. 11 depicts the on-chip size and off-chip access for parameters with respect to the group boundary on ResNet-152. It is noteworthy that the proposed scheme removes off-chip accesses for the intermediate feature-maps completely. Parameters are read once per layer by the "Main Layer". The group boundary at conv3_1 requires a minimal on-chip memory, thereby, being chosen as the boundary. To see the benefits of adding pipelined layers to the main layer, the mixed dataflow design and single layer design are implemented on FPGA. In the "single layer" design, even though shortcut data and intermediate feature-maps are access off-chip, it still consumes large on-chip resource for input buffer and partial sum buffer. As shown in Table VIII, the mixed dataflow design on ResNet-152 reduces the SRAM size and off-chip access significantly while adding small hardware overhead for pipelined layers in the 1st group. Finally, Table IX presents the comparison of the proposed mixed data flow design with previous works over the ResNet-152. The proposed design completely removes off-chip access for feature-maps, and read parameters once. Therefore, the memory bandwidth is significantly smaller than previous works. Moreover, it consumes least BRAM resources demonstrating that the proposed design provides the best performance while requiring modest off-chip memory access and on-chip memory size.

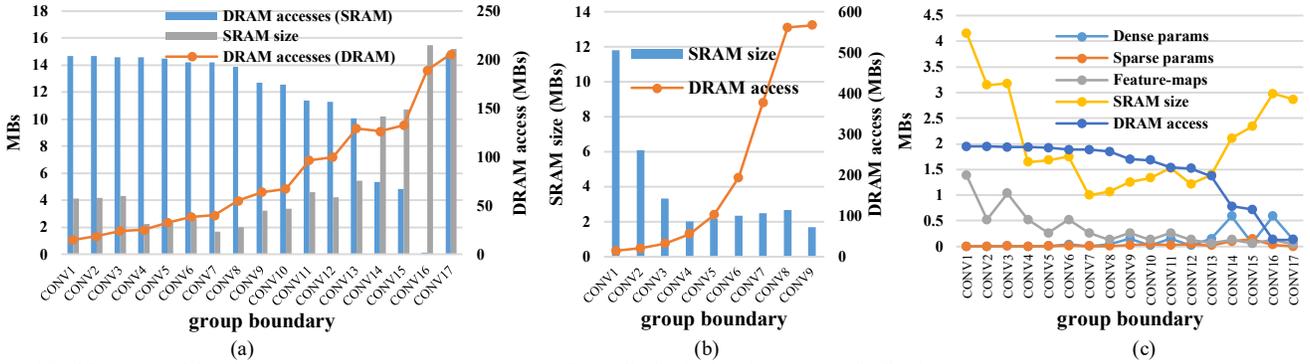

Fig. 10. SRAM size/DRAM access w.r.t group boundary. (a) 8-bit Sim-YOLO-v2. (b) 16-bit Tiny-YOLO-v2. (c) mixed precision quantization + mixed data flow scheme for Sim-YOLO-v2.

TABLE VII
TINY-YOLOV2 PERFORMANCE COMPARISON OF MIXED DATA FLOW
SCHEME AND DNNBUILDER [11]

| Features | Mixed data flow | DNNBuilder |
|---|---|---|
| FPGA board | ZC706 FPGA board (28nm) | |
| LUTs | 77.6K (218.6K) | 86.1K (218.6K) |
| FFs | 47.8K (437.2K) | 48.9K (437.2K) |
| DSPs | 892 (900) | 680 (900) |
| Block RAMs (36 Kb) | 207 (545) | 333 (545) |
| Precision | 16-bit | 16-bit |
| Frame rate (1280x384) | 21.97 | 22.1 |
| DRAM bandwidth | 1206 MB/s | 12524 MB/s |

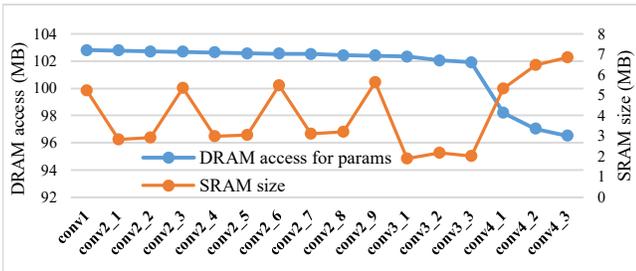

Fig. 11. The SRAM size and DRAM access w.r.t group boundaries for ResNet -152.

Finally, the additional experiment on YOLOv3 [21] is conducted. YOLOv3 has a multi-scale/multi-branch architecture and feature concatenation of different scales. Table X compares the performance of the proposed scheme against the unified design. Similar to running ResNet-152, the proposed mixed data flow design achieves higher throughput while requiring 1.6 times less BRAMs and completely eliminating off-chip access to intermediate data. These results also demonstrate that the proposed method can be extended to other detectors. It should be noted that almost all detectors (*i.e.*, YOLOv2, YOLOv3, RetinaNet, SSD, RCNN, etc.) produce the detection results, bounding boxes and classification results, on downscaled feature-maps. Hence, the mixed dataflow architecture still can be applied to these detectors. In detail, some shallow layers that work with large size feature-maps can reduce off-chip access by applying the row-reuse scheme, while deep layers that work with smaller size feature-maps can be processed by the main layer.

### D. Accelerator Design with Mixed Precision and Mixed Data Flow

The last experiment combines the mixed precision design and the mixed data flow to further reduce the on-chip memory utilization and off-chip access. It should be noted that the dense + sparse parameters of the first group are stored in BRAMs. Fig. 10(c) shows the SRAM size and DRAM access of the proposed mixed design. The group boundary between CONV7 and CONV13 reduces the SRAM size to less than 1.5 MB while requiring less than 2 MB of the DRAM access. CONV12 is chosen as the boundary to guarantee a balanced pipeline between the two groups to achieve the highest throughput. Section VI-C discusses about the advantages of the mixed data flow design to the single layer design. The combination with the mixed precision scheme further reduces the off-chip access. As shown in Table XI, the combined scheme achieves 3.2 times higher throughput while requiring 2 times less BRAMs size and 12.95 times less off-chip accesses. A single data rate (SDR) SDRAM is sufficient to support real-time performance.

TABLE VIII
RESNET152 - PERFORMANCE COMPARISON OF SINGLE LAYER DESIGN VS MIXED DATA FLOW DESIGN

| Features | Single layer | Mixed data flow |
|---|---|---|
| FPGA board | VC707 (28nm) | VC707 (28nm) |
| Frequency | 200 MHz | 200 MHz |
| LUTs | 242.4K (303.6K) | 280.4K (303.6K) |
| FFs | 188.6K (607.2K) | 220.6K (607.2K) |
| DSPs | 2112 (2800) | 2515 (2800) |
| Block RAMs (36 Kb) | 1605.5 (1030) | 715.5 (1030) |
| Precision | 16-bit | 16-bit |
| Frame rate (224x224) | 29.8 fps | 32.1 fps |
| Throughput | 675.0 GOPS | 726.0 GOPS |
| DSP efficiency | 79.9% | 72.2% |
| Off-chip FMs/frame | 122.2 MB | 0 |

TABLE IX
RESNET152 - PERFORMANCE COMPARISON TO PREVIOUS WORKS

| Features | TVLSI'18 [16] | HPCA'19 [19] | Mixed data flow |
|---|---|---|---|
| FPGA board | Arria 10 (20nm) | VC707 (28nm) | VC707 (28nm) |
| Frequency | 200 MHz | 150 MHz | 200 MHz |
| Logics | 55% | 86% | 92% |
| DSPs | 100% | 100% | 89.8% |
| Block RAMs | 87% | 99% | 69.4% |
| Precision | 16-bit | 16-bit | 16-bit |
| Frame rate | 31.3 | 26.9 | 32.1 |
| Throughput | 707.2 GOPS | 608.3 GOPS | 726.0 GOPS |
| DSP efficiency | 56.4% | 72.4% | 72.2% |
| Weight Load | Multiple times | Multiple times | Once |
| Off-chip FMs | 122.2 MB | 62.93 MB | 0 |

TABLE X
YOLOv3 - PERFORMANCE COMPARISON OF UNIFIED DESIGN VS MIXED DATA FLOW DESIGN

| Features | Unified design | Mixed data flow |
|---|---|---|
| FPGA board | VC707 (28nm) | VC707 (28nm) |
| Frequency | 200 MHz | 200 MHz |
| LUTs | 206.7K (303.6K) | 230.5K (303.6K) |
| FFs | 208.6K (607.2K) | 223.0K (607.2K) |
| DSPs | 2176 (2800) | 2640 (2800) |
| Block RAMs (36 Kb) | 1544.5 (1030) | 972.5 (1030) |
| Precision | 8-bit | 8-bit |
| Frame rate (416x416) | 10.70 fps | 11.66 fps |
| Throughput | 704.0 GOPS | 767.3 GOPS |
| DSP efficiency | 80.9% | 72.7% |
| Off-chip FMs/frame | 112.1 MB | 0 |

TABLE XI
SIMYOLOV2 - PERFORMANCE COMPARISON OF SINGLE LAYER DESIGN AND MIXED PRECISION + MIXED DATA FLOW

| Features | Single layer design | Mixed precision + mixed data flow |
|---|---|---|
| FPGA board | VC707 | VC707 |
| Frequency | 200 MHz | 200 MHz |
| LUTs | 138.8K (303.6K) | 154.4K (303.6K) |
| FFs | 96.1K (607.2K) | 97.6K (607.2K) |
| DSPs | 1056 (2800) | 587 (2800) |
| Block RAMs (36 Kb) | 639 (1030) | 314.5 (1030) |
| Precision | 8-bit | Mixed precision |
| Frame rate (416x416) | 22.58 fps | 72.11 fps |
| Throughput | 388 GOPS | 1239 GOPS |
| Off-chip FMs/frame | 14.58 MB | 0 |
| Off-chip access/frame | 29.79 MB | 2.30 MB |

*E. Related Works*

Research in [50] presents a heterogeneous weight quantization including both equal-distance and mixed powers-of-two methods for different layers of Tiny-YOLO v2. However, the compression causes 2.7% accuracy loss compared to the original model. Another research in [51] combines both quantization and pruning to accelerate CNN training process. It presents an architecture to utilize both inference and back-propagation sparsity to achieve low operation complexity. Different from this work, the proposed work aims to reduce off-chip access/on-chip memory size for inference only.

To reduce off-chip access for intermediate data, the accelerator designs in [60] and [61] perform the pipelined computation between layers. The ASIC design in [60] utilizes 5-bit look-up-table to remove large hardware cost of multiplications. As reported in [60], the accuracy loss of this approach is significant when the CNNs become deeper. The study in [61] optimizes the CNN structure using depth-wise convolution [53] so that the entire weights are able to be stored in BRAMs of a FPGA chip owing to lightweight and shallow CNN structure. However, these works would not scale up well for larger datasets and deeper networks that are well supported by the proposed scheme.

The study in [58] introduces an integrated CNN accelerator design with a dynamic fixed-point quantization strategy to minimize the computational loss while saving hardware resources and memory bandwidth. Another work in [59] proposes a CNN hardware design which supports configurable multi-precision computation using single bit RRAM. In this design, each layer is computed using a different number of bits, which can significantly reduce energy consumption. Different from these studies, this paper proposes a coarse-grained intra-layer mixed precision quantization scheme and the corresponding hardware design, and evaluates high compatibility of the proposed method with various network structures and various datasets.

Regarding the CNN compression, there are many previous works aim to reduce model size of CNNs [25], [39], [41], [42], [52]-[54], [62]. Binary [39] and ternary [54] weight quantization schemes compress the network by 32 and 16 times, respectively, but cause a significant accuracy loss on ImageNet dataset. Pruning [25] reduces parameter size of convolutional layers of VGGNet by 4.5 times with no loss of accuracy. Another hardware-aware technique in [62] also presents a constrained pruning approach to achieve a similar pruning ratio as [25] while balancing the computation for the sparse CNN accelerators. However, it should be noted that the experimental results in Section VI-A show that the proposed scheme achieves significantly higher compression ratio than the pruning approach with a same accuracy. Moreover, the workload imbalance is also solved by a channel-wise quantization in the proposed scheme. Meanwhile, an object detector made by compact network such as MobileNetv2 and SSDLite [53] outperforms YOLO v2 on COCO dataset (22.1% vs 21.6%) while keeping the network size 10 times smaller. Whereas, the proposed mixed precision compresses networks by 22.66 – 28.93 times on PASCAL VOC, COCO, and ImageNet datasets with negligible loss.

Finally, AutoML based methods such as HAQ [41] and RaQu [30] proposed a fined-grained inter-layer mixed precision quantization, where the computation within a layer uses a same number of bits. Different from these works, the proposed scheme employs a coarse-grained intra-layer mixed precision using Bayesian approach. Therefore, the AutoML approaches from [41] and [30] can also be applied to the proposed quantization to manipulate the weight ratios within the layer.

## VII. CONCLUSION

This paper proposes a layer-specific design that employs different organizations that are optimized for the different layers. The proposed design employs two layer-specific optimizations: layer-specific mixed precision and layer-specific mixed data flow. The mixed precision scheme causes a negligible accuracy loss while reducing the model size significantly compared to that in a full-precision network. As a result, the proposed schemes significantly outperform the previous works in terms of both throughput, off-chip access, and on-chip memory requirement.

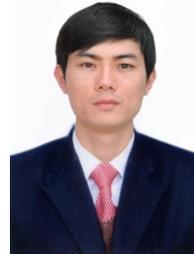

**Duy Thanh Nguyen** received the B.S. degree in electrical engineering from Hanoi University of Science and Technology, Hanoi, Vietnam, M.S. degree in Electrical and Computer Engineering from Seoul National University, Seoul, Korea, in 2011 and 2014, respectively. He is currently working toward the Ph.D. degree in Electrical and Computer Engineering at Seoul National University.

His research interests include computer architecture, memory system, SoC design for computer vision applications.

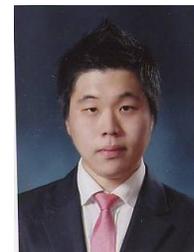

**Hyun Kim** received the B.S., M.S. and Ph.D. degrees in Electrical Engineering and Computer Science from Seoul National University, Seoul, Korea, in 2009, 2011 and 2015, respectively. From 2015 to 2018, he was with the BK21 Creative Research Engineer Development for IT, Seoul National University, Seoul, Korea, as a BK Assistant Professor. In 2018, he joined the Department of Electrical and Information Engineering, Seoul National University of Science and Technology, Seoul, Korea, where he is currently working as an Assistant Professor. His research interests are the areas of algorithm, computer architecture, memory, and SoC design for low-complexity multimedia applications and deep neural networks.

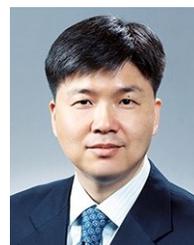

**Hyuk-Jae Lee** received the B.S. and M.S. degrees in electronics engineering from Seoul National University, South Korea, in 1987 and 1989, respectively, and the Ph.D. degree in Electrical and Computer Engineering from Purdue University, West Lafayette, IN, in 1996. From 1996 to 1998, he was with the Faculty of the Department of Computer Science, Louisiana Tech University, Ruston, LS. From 1998 to 2001, he was with the Server and Workstation Chipset Division, Intel Corporation, Hillsboro, OR, as a Senior Component Design Engineer. In 2001, he joined the School of Electrical Engineering and Computer Science, Seoul National University, South Korea, where he is currently a Professor. He is a Founder of Mamurian Design, Inc., a fabless SoC design house for multimedia applications. His research interests are in the areas of computer architecture and SoC design for multimedia applications.